\documentclass[prodmode,acmtecs]{acmsmall}
\usepackage[ruled]{algorithm2e}
\renewcommand{\algorithmcfname}{ALGORITHM}
\SetAlFnt{\small}
\SetAlCapFnt{\small}
\SetAlCapNameFnt{\small}
\SetAlCapHSkip{0pt}
\IncMargin{-\parindent}

\acmYear{2014}

\usepackage{amsfonts}
\usepackage{times}
\usepackage{subfigure}
\usepackage{hyperref}
\usepackage{graphicx}
\usepackage{multirow}
\usepackage{paralist}
\usepackage{url}
\usepackage{amssymb}
\begin{document}

\markboth{Q. Cui et al.}{KNET: A General Framework for Learning Word Embedding using Morphological Knowledge}

\title{KNET: A General Framework for Learning Word Embedding using Morphological Knowledge}

\author{
Qing Cui
\affil{Tsinghua University}
Bin Gao
\affil{Microsoft Research}
Jiang Bian
\affil{Microsoft Research}
Siyu Qiu
\affil{Nankai University}
Tie-Yan Liu
\affil{Microsoft Research}
}

\begin{abstract}
  Neural network techniques are widely applied to obtain high-quality distributed representations of words, i.e., word embeddings, to address text mining, information retrieval, and natural language processing tasks. Recently, efficient methods have been proposed to learn word embeddings from context that captures both semantic and syntactic relationships between words. However, it is challenging to handle unseen words or rare words with insufficient context. In this paper, inspired by the study on word recognition process in cognitive psychology, we propose to take advantage of seemingly less obvious but essentially important morphological knowledge to address these challenges. In particular, we introduce a novel neural network architecture called KNET that leverages both contextual information and morphological word similarity built based on morphological knowledge to learn word embeddings. Meanwhile, the learning architecture is also able to refine the pre-defined morphological knowledge and obtain more accurate word similarity. Experiments on an analogical reasoning task and a word similarity task both demonstrate that the proposed KNET framework can greatly enhance the effectiveness of word embeddings.
\end{abstract}

\acmformat{Qing Cui, Bin Gao, Jiang Bian, Siyu Qiu, Tie-Yan Liu, 2014. Learning Effective Word Embedding using Morphological Knowledge.}

\begin{bottomstuff}
Author's addresses: Qing Cui, Department of Mathematical Sciences, Tsinghua University, Beijing, 100084, P. R. China; Bin Gao, Jiang Bian, and Tie-Yan Liu, Microsoft Research, 13F, Bldg 2, No. 5, Danling St, Beijing, 100080, P. R. China; Siyu Qiu, Nankai University, Tianjin, 300071, P. R. China.
\end{bottomstuff}

\maketitle

\section{Introduction}\label{sec-introduction}

Neural network techniques have been widely applied to solve text mining, information retrieval (IR), and natural language processing (NLP) tasks, the basis of which yields obtaining high-quality distributed representations of words, i.e., word embeddings.  In recent years, efficient methods, such as the continuous bag-of-word (CBOW) model and the continuous Skip-gram (Skip-gram) model \cite{Mikolov2013nips}, have been proposed to leverage the surrounding context of a word in documents to transform words into vectors (i.e., word embeddings) in a continuous space, which surprisingly captures both semantic and syntactic relationships between words. The underlying principle in these works lies in that words that are syntactically or semantically similar should have similar surrounding contexts.

While the aforementioned works have demonstrated their effectiveness in various tasks, they also suffer from a couple of limitations.
\begin{enumerate}
  \item It is difficult to obtain word embeddings for new words since they are not included in the previous vocabulary. Some previous studies \cite{Mikolov2012phd} used a default index to represent all unknown words, but such a solution will inevitably lose information for emerging words.
  \item The embeddings for rare words are unreliable due to the insufficient surrounding contexts. Since the aforementioned works adopt statistical methods, when a word has only a few occurrences in the training data, they will fail in extracting statistical clues to correctly map the word into the embedding space.
\end{enumerate}

In sharp contrast, according to the studies on word recognition in cognitive psychology \cite{ehri1991development,ehri2005learning}, when a human looks at a word, no matter new or rare, she can figure out effective ways to understand it. For instance, one sometimes conducts phonological recoding through blending graphemes into phonemes and blend syllabic units into recognizable words; one may also analyze the root/affix of the new word so as to build its connections with her known words. Suppose the new word is \emph{inconveniently}. Given its root and affixes, i.e., \emph{in-convenient-ly}, it is natural to guess that it is the adverb form of \emph{inconvenient} and the latter is probably the antonym of \emph{convenient}. Henceforth, morphological word similarity can act as an effective bridge for understanding new or rare words based on known words in the vocabulary. Inspired by this word recognition process, we propose using morphological knowledge to enhance the deep learning framework for learning word embedding. In particular, beyond the contextual information already used in CBOW and Skip-gram, we take advantage of morphological similarity between words in the learning process so as to handle new or rare words.

Although the morphological knowledge contains invaluable information, it might be risky to blindly rely on it. The reason is that the prediction based on morphological word similarity is somehow only a kind of guess, and there exists many counter examples inconsistent with it. For example, if only looking at the morphological similarity, one may link \emph{convention} to \emph{convenient} since they share a long substring. However, it is clear that these two words are neither syntactically nor semantically similar. In this case, if we stick to the morphological knowledge, the effectiveness of the learned word embeddings could be even worse. To tackle this issue, we once again leverage the findings regarding word recognition in cognitive psychology \cite{ehri1991development,ehri2005learning}. It has been revealed that humans can take advantage of the contextual information (both the context at the reading time and the context in her memory) to correct the unreliable morphological word similarity. By comparing their respective contexts, one can distinguish between \emph{convenient} and \emph{convention} and weaken the morphological connection between these two words in her mind. Inspired by this, we also propose updating the morphological knowledge during our learning process. Specifically, we will not fully trust the morphological knowledge, and will change it so as to maximize the consistency between contextual information and morphological word similarity.

To sum up the discussions above, we actually develop a novel neural network architecture that can leverage morphological word similarity for word embedding. Our proposed framework consists of a contextual information branch and a morphological knowledge branch. On one hand, we adopt the state-of-the-art Skip-gram model \cite{Mikolov2013nips} as our contextual information branch for its efficiency and effectiveness. On the other hand, we explore edit distance, longest common substring similarity, morpheme similarity, and syllable similarity as morphological knowledge to build a relation matrix between words, and put the relation matrix into the morphological knowledge branch. These two branches share the same word embedding space, and they are combined together using tradeoff coefficients in order to feed forward to the output layer to predict the target word. The back propagation stage will modify the tradeoff coefficients, word embeddings, and the weights in the relation matrix layer by layer. We call the proposed framework as \emph{KNET}, for it is a \emph{K}nowledge-powered neural \emph{NET}work. We have conducted experiments on a publicly available dataset, and the results demonstrate that our proposed KNET method can help produce improved word representations as compared with the state-of-the-art methods on an analogical reasoning task and a word similarity task.

The main contributions of the paper include:
\begin{enumerate}
  \item We have proposed a general and robust neural network framework called KNET that can effectively leverage both contextual information and morphological knowledge to learn word embeddings.
  \item The KNET framework can learn high quality word embeddings especially on rare words and new words with the help of morphological knowledge even when the knowledge is not very reliable.
  \item We also conduct some experimental study to gain insight about how KNET can benefit from noisy knowledge and balance between contextual information and morphological knowledge.
\end{enumerate}

The rest of the paper is organized as follows. We briefly review the related work on word embedding using deep neural networks in Section \ref{sec-related}. In Section \ref{sec-algorithm}, we describe the proposed framework to leverage morphological knowledge in word embedding using deep neural networks. The experimental results are reported in Section \ref{sec-experiment}. The paper is concluded in Section \ref{sec-conclusion}.

\section{Related Work}\label{sec-related}

Word embedding as continuous vectors has been studied for a long time \cite{Hinton1986dr}. Many different types of models were proposed for learning continuous representations of words, such as the well-known Latent Semantic Analysis (LSA)~\cite{hofmann:plsa} and Latent Dirichlet Allocation (LDA)~\cite{blei:lda}. However, such probabilistic approaches usually yield the limitation in terms of scalability. Recently, deep learning methods have been applied to obtain continuous word embeddings to solve a variety of text mining, information retrieval, and natural language processing tasks \cite{Collobert2008,Glorot2011,Mikolov2013w2v,Mikolov2013nips,Socher2011RNN,Turney:arXiv1310.5042,Turney2010,DengHG13,Collobert2011,MnihH08,TurianRB10}. For example, Collobert \emph{et al} \cite{Collobert2008,Collobert2011} proposed a unified neural network architecture that learns word representations based on large amounts of unlabeled training data, to deal with several different natural language processing tasks.

Most recently, Mikolov \emph{et al} \cite{Mikolov2013w2v,Mikolov2013nips} proposed the continuous bag-of-words model (CBOW) and the continuous skip-gram model (Skip-gram) for learning distributed representations of words also from large amount of unlabeled text data; these models can map the semantically or syntactically similar words to close positions in the word embedding space, based on the intuition that the contexts of the similar words are similar. In particular, in the Skip-gram model, a sliding window is employed on the input text stream to generate the training samples. In each sliding window, the model tries to use the central word as input to predict the surrounding words. Specifically, the input word is represented in the 1-of-$V$ format, where $V$ is the size of the vocabulary of the training data and each word in the vocabulary is represented as a vector with only one non-zero element. In the feed-forward process, the input word is first mapped into the embedding space by the weight matrix $M$. After that, the embedding vector is mapped back to the 1-of-$V$ space by another weight matrix $M'$, and the resulting vector is used to predict the surrounding words after applying \emph{softmax} function on it. In the back-propagation process, the prediction errors are propagated back to the network to update the two weight matrices. When the training process converges, the weight matrix $M$ is used as the learned word embeddings. Though the above works like Skip-gram perform good on some NLP tasks, they still cannot produce high-quality word embeddings for rare words and unknown words since they do not leverage the rich extra knowledge when learning word embeddings.

There are some knowledge related word embedding works in the literature, but most of them were targeted at the problems of knowledge base completion and enhancement \cite{bordes2011learning,socher2013reasoning,weston2013connecting} rather than producing high-quality word embeddings, which is different with our work. In contrast, some recent efforts have explored how to take advantage of knowledge to product better word embedding. For example, Qiu et al.~\cite{qiu2014coling} introduced a co-learning framework to produce both the word representation and the morpheme representation such that each of them can be mutually reinforced. Yu et al.~\cite{Yu:2014} proposed a new learning objective that integrates both a neural language model objective and a semantic prior knowledge objective which can result in better word embedding for semantic tasks. Moreover, a recent work~\cite{bian2014knn} took empirical studies on how to incorporate various types of knowledge in order to enhance word embedding. According to this work, morphological, syntactic, and semantic knowledge are all valuable to improve the quality of word embedding. In this paper, as we aim at obtaining high-quality word embeddings for rare words and unknown words, we focus on leveraging morphological knowledge since it can generate critical correlation between rare/unknown words with popular ones.

Some previous works have attempted to include morphology in continuous models, especially in the speech recognition field, including Letter n-gram~\cite{sperr2013morph} and feature-rich DNN-LMs~\cite{mousa2013morph}. The first work improves the letter-based word representation by replacing the 1-of-$V$ word input of restricted Boltzman machine with a vector indicating all n-grams of order n and smaller that occur in the word. Additional information such as capitalization is added as well. In the model of feature-rich DNN-LMs, the authors expand the inputs of the network to be a mixture of 142 selected full words and morphemes together with their features such as morphological tags. Both of these works intend to capture more morphological information so as to better generalize to rare/unknown words and to lower the out-of-vocabulary rate.

In the NLP and text mining domain, Luong \emph{et al} \cite{luong2013better} proposed a morphological Recursive Neural Network (morphoRNN) that combines recursive neural networks and neural language models to learn better word representations, in which they regarded each morpheme as a basic unit and leveraged neural language models to consider contextual information in learning morphologically-aware word representations. We will compare our proposed model with morphoRNN in Section \ref{sec:morphoRNN}.

\section{Word Embedding Powered by Morphological Knowledge}\label{sec-algorithm}

We first introduce how people learn words and understand text by leveraging the morphological knowledge, and then describe the knowledge-powered neural network architecture for learning effective word embedding based on both contextual information and morphological knowledge. Afterwards, we mention four types of morphological knowledge that are often used by people as well as our framework.

\subsection{Word Recognition Process}\label{sec-human}

According to the study on word recognition in cognitive psychology \cite{ehri1991development,ehri2005learning}, when human learns a new language, she usually starts from learning some basic words and gradually enlarges her vocabulary during the learning process. She also learns the language grammars and morphological knowledge so as to build cross links between words in her knowledge base, e.g., the adjective form of \emph{care} is \emph{careful} and its adverb form is \emph{carefully}. When she encounters an unknown or unfamiliar word, she will try to explore several different channels to recognize it \cite{ehri1991development}:

\textbf{Recoding (or Decoding).} One can either sound out and blend graphemes into phonemes, or work with larger chunks of letters to blend syllables into recognizable words. For example, \emph{psychology} can be pronounced as \emph{psy-cho-lo-gy}, in which \emph{psy} means know or study, \emph{cho} means mind or soul, and \emph{logy} means academic discipline. Thus, she may guess \emph{psychology} is an academic discipline that studies something in the mind or soul.

\textbf{Analogizing.} One can use her known words to read the new word \cite{goswami1986children}. If the new word is morphologically similar to several known words, she will guess the meaning of the new words based on the meanings of these known words. For example, \emph{admob} appears in a news article as a new word to a reader. The reader may quickly understand that it is related the advertisements on mobile devices, simply because \emph{admob} is composed by \emph{ad} and \emph{mob}, which are substrings of \emph{advertisement} and \emph{mobile}, respectively.

\textbf{Prediction.} One can use context and letter clues to directly guess the meaning of the unknown word \cite{chapman1998language}. Sometimes, one may even retrieve the context of the word in her memory and make associations to the current context to guess the meaning of the word. For example, \emph{inmate} is an unknown word to a reader, but according to the context \emph{Inmates and police officers held a basketball game in the Fox River Prison last Tuesday evening}, she can easily guess that \emph{inmate} means prisoner in the sentence.

In the above process, the different channels may reinforce each other. On one hand, sometimes contexts could be insufficient, e.g., there are simply not many contexts surrounding the unknown word, and there are no historical context in the memory either. In this case, it is extremely hard to directly predict the meaning of the word. In contrast, decoding and analogizing could do a good job since they can work in a context-free manner. On the other hand, sometimes decoding and analogizing can result in errors. For example, \emph{convention} and \emph{convenient} are morphologically very similar since they share a long substring \emph{conven}; however, their meanings are quite different. In this case, blindly relying on morphological knowledge will bring in a lot of noises, but contextual information can help one to successfully distinguish these two words. By refining her morphological knowledge with the help of the contextual information, one can avoid the misrecognition.

Please note that all the above process happens within just a second, which enables human to be super powerful in recognizing unknown or unfamiliar words. This phenomenon strongly inspires us to leverage both morphological knowledge and contextual information to learn word embeddings.  Accordingly, we propose a novel neural network architecture that consists of a morphological knowledge branch and a contextual information branch. Details will be given in the next subsection.

\subsection{Neural Network Architecture for KNET}\label{sec-model}

In this subsection, we describe our proposed new neural network architecture that leverages both contextual information and morphological knowledge to learn word embedding. We use the Skip-gram model \cite{Mikolov2013nips} as the basis of our proposed framework.\footnote{Note that although we task the Skip-gram model as an example to illustrate our framework, the similar framework can be developed on the basis of any other word embedding models.} Skip-gram is a neural network model to learn word representations, the underlying principle of which is that similar words should have similar contexts.

To be more specific, given a sequence of training words $w_1, w_2, \dots, w_T$, the objective of the Skip-gram model is to maximize the following average log probability,
\begin{equation}\label{eq:obj}
\frac{1}{T}\sum_{t=1}^T{\sum_{-N\le j\le N,j\ne 0}{\log p(w_{t+j}|w_t)}},
\end{equation}
where $w_I$ denotes the input word (i.e., $w_t$), $w_O$ denotes the output word (i.e., $w_{t+j}$), and $N$ indicates the size of the sliding window is $2N + 1$. The conditional probability $p(w_{t+j}|w_t)$ is defined using the following \emph{softmax} function,
\begin{equation}\label{eq:softmax}
p(w_O|w_I)=\frac{\exp({v^{'}_{w_O}}^{T} v_{w_I})}{\sum_{w}\exp({v^{'}_{w}}^{T} v_{w_I})},
\end{equation}
where $v_w$ and $v_{w}^{'}$ are the \emph{input} and \emph{output} representation vectors of $w$, and the sum in the denominator is over all words in the vocabulary.

It is difficult and impractical to directly optimize the above objective because computing the derivative is proportional to the vocabulary size, which is often very large. Several approaches \cite{HierarchicalSoftmax,Bengio03ImportanceSampling,Bengio08ImportanceSampling} have been employed to tackle this problem. The state-of-the-art method is noise-contrastive estimation (NCE) \cite{Gutmann2012NCE}, which aims at fitting unnormalized probabilistic models. NCE can approximate the log probability of the \emph{softmax} function by performing logistic regression to discriminate between the observed data and some artificially generated noises. It was first adapted in the neural language model in \cite{MnihICML2012NCE}, and was then applied to the inverse vector log-bilinear model \cite{MnihNIPS2013}. Another simpler method is negative sampling (NEG) \cite{Mikolov2013nips}, which generates $k$ noise samples for each input word to estimate the objective.

By using NEG, the \emph{softmax} conditional probability $p(w_{t+j}|w_t)$ will be replaced by
\begin{equation}\label{eq:NEG}
J(\theta)=\log\sigma({v^{'}_{w_O}}^{T} v_{w_I}) + \sum_{i=1}^{k}\mathbb{E}_{{w_i}\sim P_n(w)}\left[\log \sigma(-{v^{'}_{w_i}}^{T} v_{w_I}))\right],
\end{equation}
where $\theta$ is the model parameter including the word embeddings, $\sigma$ denotes the logistic function, and $P_n(w)$ represents the noise distribution which is set as the 3/4 power of the unigram distribution $U(w)$, i.e., $P_n(w) = U(w)^{3/4}/Z$ ($Z$ is a normalizer) \cite{Mikolov2013nips}. Then, we can estimate the gradient of $J(\theta)$ by computing
\begin{equation}\label{eq:gradient}
\frac{\partial J(\theta)}{\partial\theta} = (1-\sigma({v^{'}_{w_O}}^{T} v_{w_I}))\frac{\partial{v^{'}_{w_O}}^{T} v_{w_I}}{\partial\theta} - \sum_{i=1}^{k}\left[\sigma({v^{'}_{w_i}}^{T} v_{w_I})\frac{\partial{v^{'}_{w_i}}^{T} v_{w_I}}{\partial\theta}\right].
\end{equation}

By summing over $k$ noise samples instead of a sum over the entire vocabulary, the training time yields linear scale to the number of noise samples and becomes independent of the vocabulary size.

\begin{figure}[ht]
\vskip 0in
\begin{center}\vspace{0pt}
\centerline{\includegraphics[width=0.9\columnwidth]{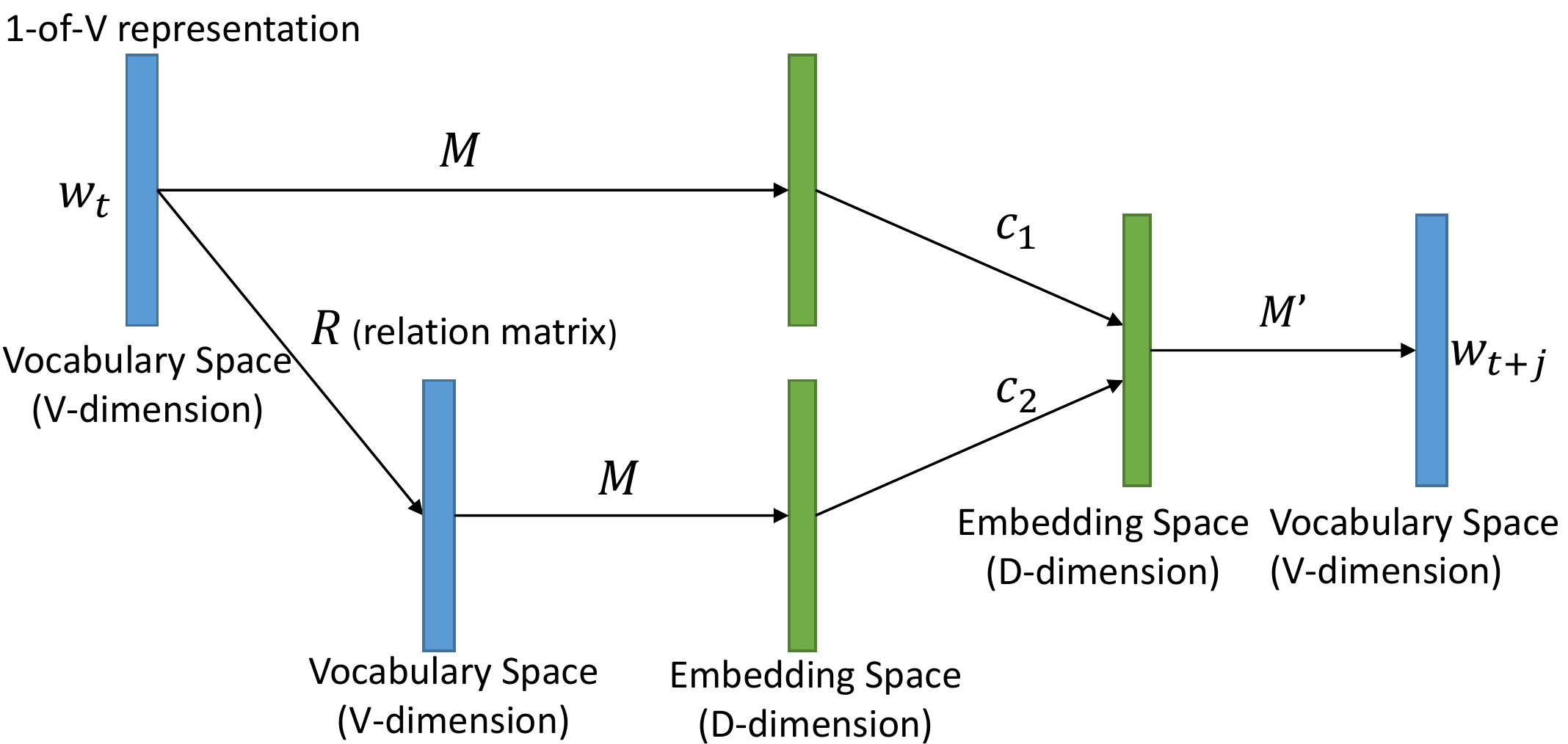}}
\caption{The neural network architecture of the proposed KNET framework} \vspace{0pt}
\label{figure-NN}
\end{center}
\vskip 0in
\end{figure}

To incorporate morphological knowledge into the learning process, we propose a new neural network architecture. Beyond the basic Skip-gram model that predicts a target word based on its context, the proposed new method introduces a parallel branch that leverages morphological knowledge to assist predicting target word, as shown in Figure~\ref{figure-NN}. Intuitively, when a word $w_t$ is the central word in the context window, we predict the surrounding words by leveraging not only the representation of word $w_t$ as contextual information (referred as \emph{contextual information branch}) but also the representations of the words that are morphologically similar to $w_t$ (referred as \emph{morphological knowledge branch}). Therefore, the objective of the proposed model is the same as (\ref{eq:obj}) (i.e., we want to maximize the average probability of word prediction) except that we replace the input word representation $v_{w_I}$ in the softmax function (\ref{eq:softmax}) by a new formulation which is combined from both the contextual information branch and the morphological knowledge branch. We introduce the detailed formulation of $v_{w_I}$ as below.

According to Figure \ref{figure-NN}, to obtain the representation of a central word $w_t$ from the morphological knowledge branch, it is necessary to find the set of words that are morphologically similar to $w_t$, which is denoted as $R_t$. Then, we can extract the embedding of each word in $R_t$ from the embedding matrix $M$ shared with the contextual information branch. After that, a corresponding knowledge representation of $R_t$ can be computed by feeding forward the relationship layer, which is written as
\begin{equation}\label{eq:vRt}
v_{R_t}=\sum_{w\in R_t}s(w_t, w)v_w,
\end{equation}
where $s(w_t, w)$ is the similarity score, the methods of computing which will be introduced in Section \ref{sec-knowledge}. Actually $v_w$ is the $i$-th row of matrix $M$ where $i$ is the index of the word $w$ in the vocabulary, and $s(w_1, w_2)$ is the element of relation matrix $R$ at $(i, j)$ which are the indices of words $w_1$ and $w_2$ respectively. To ensure the quality of morphological knowledge and control the number of parameters, we only leverage the top words with highest morphological similarity scores as $R_t$. For example, in our experiments, an input word can only connect to at most 5 words in the relationship layer. This sparse structure will not change during training, and only the weights of these connections will be updated. Therefore, we will not suffer from a huge number of parameters even if $R$ is learned.

Finally, an aggregated representation of the input word, denoted as $v_{w_I}$, can be calculated as the weighted sum of the representations from the contextual information branch and the morphological knowledge branch, i.e.,
\begin{equation}
v_{w_I}=c_1(w_t)v_{w_t}+c_2(w_t)v_{R_t},\label{eq:wI}
\end{equation}
where $c_1(\cdot)$ and $c_2(\cdot)$ are the functions of $w_t$ and yield much dependency on the word frequency. Intuitively, frequent words are associated with much more training samples than rare words, such that it is easy to collect rich contextual information for frequent words, while the contextual information for rare words might be insufficient. In contrast, the volume of morphological knowledge of a word usually has little correlation to the word frequency, and thus rare words can still rely more on the morphological knowledge even though the contextual information is not reliable. Therefore, the balancing function $c_1(\cdot)$ and $c_2(\cdot)$ should be related to word frequency. Specifically, we divide the words into a number of buckets according to their frequencies, and all the words in the same bucket will share the same values of $c_1(\cdot)$ and $c_2(\cdot)$.

A more explicitly intuitive way to interpret the above model is as follows. For each word $w_t$, we use one row in the embedding matrix $M$ to encode its contextual embedding. In addition, by using matrix $R$, we can identify a couple of morphologically similar words to $w_t$. Then we can also extract the contextual embeddings of these similar words from $M$ and take the weighted average of these embedding vectors as the morphological embedding for the original word $w_t$. Then finally the overall embedding of $w_t$ is computed as the weighted combination of its contextual embedding and morphological embedding. Matrix $M'$ is used to predict the surrounding word $w_{t+j}$ based on the overall embedding of $w_t$. In the back-propagation process, the parameters in $M$, $R$, $M^{'}$, and multiple pairs of $c_1$ and $c_2$ (corresponding to different frequency buckets) are updated. When the training process converges, we take the matrix $M$ as the learned word embeddings. In our implementation, we take the NEG strategy to calculate the gradient in (\ref{eq:gradient}), in which $v_{w_I}$ is substituted by (\ref{eq:wI}), and learn the parameters with standard gradient descent techniques. We call the proposed framework as \emph{KNET}, considering that it is a \emph{K}nowledge-powered neural \emph{NET}work.

\subsection{Morphological Knowledge}\label{sec-knowledge}

As compared to Skip-gram, the uniqueness of our model lies in the introduction of the morphological knowledge branch. In this subsection, we will make discussions on how we realize this new branch. In particular, we propose four types of naturally defined morphological knowledge. Note that this is not a complete study on morphological knowledge, but we can use these four specific types as examples to show the effectiveness of the proposed framework. Any other types of morphological knowledge can be used under the KNET framework.

\subsubsection{Edit Distance Similarity (Edit)}

Edit distance is a way of quantifying how dissimilar two strings (e.g., words) are by counting the minimum number of operations required to transform one string into the other. The operations might be \emph{letter insertion}, \emph{letter deletion}, or \emph{letter substitution}. We calculate the edit distance similarity score for two words $w_1$ and $w_2$ as
$$s_{Edit}(w_1, w_2)=1-\frac{d(w_1, w_2)}{\max(l(w_1), l(w_2))},$$
where $d(w_1,w_2)$ represents the edit distance of the two words and $l(w_1), l(w_2)$ are the corresponding word lengths.

\subsubsection{Longest Common Substring Similarity (LCS)}

Longest common substring similarity is defined as the ratio of the length of the longest shared substring of two words (denoted by $g(w_1,w_2)$ ) and the length of the longer word, i.e.,
$$s_{LCS}(w_1, w_2)=\frac{g(w_1, w_2)}{\max(l(w_1), l(w_2))}.$$

\subsubsection{Morpheme Similarity (Morpheme)}

Morpheme similarity is calculated based on the shared roots (or stems) and affixes (prefix and suffix) of two words. Suppose each word of $w_1$ and $w_2$ can be split into a set of morphemes (denoted by $F(w_1)$ and $F(w_2)$), then the morpheme similarity of the two words is calculated as
$$s_{Morpheme}=\frac{ |F(w_1) \bigcap F(w_2)|}{\max(|F(w_1)|, | F(w_2)|)},$$
where $|\cdot|$ outputs the size of the set.

\subsubsection{Syllable Similarity (Syllable)}

Syllable similarity is calculated based on the shared syllables of two words. Suppose both $w_1$ and $w_2$ can be split into a set of syllables (denoted by $G(w_1)$ and $G(w_2)$), then the syllable similarity of the two words is calculated as
$$s_{Syllable}=\frac{ |G(w_1) \bigcap G(w_2)|}{\max(|G(w_1)|, |G(w_2)|)}.$$

In addition to using these four types of morphological word similarity separately, one can also combine them together. In the next section, we will conduct experimental study on all these different choices.

\section{Experimental Evaluation}\label{sec-experiment}

In this section, we report the experimental results regarding the effectiveness of our proposed KNET framework. Our experiments are mainly composed of three parts. In the first part, we compare KNET with several baselines based on Skip-gram to show the effectiveness and robustness of our framework. Then we compare KNET with morphoRNN on two word similarity tasks, one mainly contains frequent words the other contains lots of rare words, to show that our framework can achieve high quality word embedding on rare and new words. After that, we conduct some case studies to gain deeper understanding about how KNET can benefit from noisy knowledge to obtain high quality word embedding on rare words; we also give an empirical study to gain insight about the balancing function between the contextual information branch and the morphological knowledge branch.

\subsection{Evaluation Tasks}
We evaluated the performance of the learned word representations on the following two tasks.

\subsubsection{Analogical Reasoning Task}
The analogical reasoning task was introduced by Mikolov \emph{et al} \cite{Mikolov2013w2v}. The task consists of 19,544 questions of the form ``$a$ is to $b$ is as $c$ is to \underline{ }\underline{ }'', denoted as $a$ : $b$ $\rightarrow$ $c$ : ?. Suppose $\overrightarrow{w}$ is the learned word representation vector of word $w$ normalized to unit norm. Following \cite{Mikolov2013w2v}, we answer this question by finding the word $d^*$ whose representation vector is the closest to vector $\overrightarrow{b} - \overrightarrow{a} + \overrightarrow{c}$ according to cosine similarity excluding $b$ and $c$, i.e.,
$$d^*=\arg\max_{x \in V, x\ne b, x\ne c}(\overrightarrow{b} - \overrightarrow{a} + \overrightarrow{c})^T\overrightarrow{x}.$$
The question is regarded as answered correctly only when $d^*$ is exactly the answer word in the evaluation set. There are two categories in the task, with 8,869 semantic analogies (e.g., \emph{England} : \emph{London} $\rightarrow$ \emph{China} : \emph{Beijing}) and 10,675 syntactic analogies (e.g., \emph{amazing} : \emph{amazingly} $\rightarrow$ \emph{unfortunate} : \emph{unfortunately}).

\subsubsection{Word Similarity Task}\label{sec:wordSimTask}
WordSim-353 \cite{Finkelstein2001wordsim353} is a standard dataset for evaluating vector space models on word similarity. It contains 353 pairs of nouns without context. Each pair is associated with 13 to 16 human judgments on similarity and relatedness on a scale from 0 to 10. For example, (\emph{cup}, \emph{drink}) received an average score of 7.25, while (\emph{cup}, \emph{substance}) received an average score of 1.92. To evaluate the quality of the learned word embeddings, we computed Spearman's $\rho$ correlation between the similarity scores calculated by word embeddings and the human judgments.

\begin{figure}[ht]
\vskip 0in
\begin{center}\vspace{0pt}
\centerline{\includegraphics[width=0.9\columnwidth]{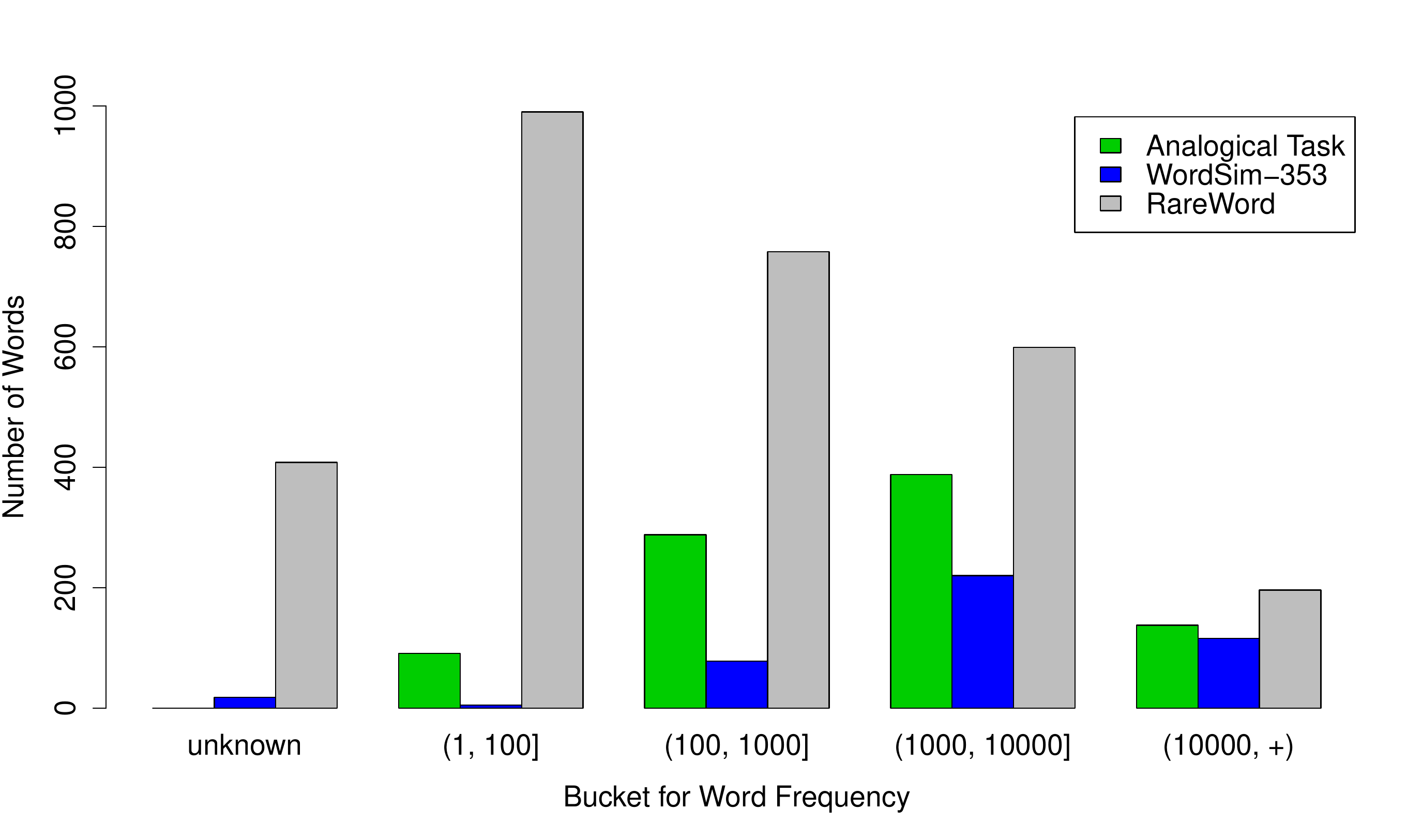}}\vspace{0pt}
\caption{Word frequency distributions for the word similarity test sets in the \emph{enwiki9} dataset} \vspace{0pt}
\label{figure-frequency}
\end{center}
\vskip 0in
\end{figure}

In addition to WordSim-353, we also used the RareWord dataset \cite{luong2013better} to test the performance of the proposed model, which contains 2,034 pairs of rare words. According to the frequency distribution in the training data \emph{enwik9} (see Figure \ref{figure-frequency}), half of the words in the RareWord are tail words (word frequency $<100$), while WordSim-353 is mainly composed of frequent words. Furthermore, the RareWord dataset contains more than 400 unknown words (which have not appeared in the training data and thus do not have word embeddings available by themselves). We make use of these unknown words to test the capability of our KNET model in dealing with new words. We use $v_{R_t}$ in (\ref{eq:vRt}) as the embedding for an unknown word. Specifically, we computed its similarity to all the known words using a certain type of morphological knowledge, and then we selected the top 5 closest known words and calculated the linear combination of their embedding vectors as the representation for the unknown word (the normalized similarity scores were used as the combination weights).

\subsection{Experimental Setup}

\subsubsection{The Construction of Relation Matrix $R$}
In our experiments, we employed four types of morphological knowledge. \emph{Edit} and \emph{LCS} can be computed directly from the definitions. For \emph{Morpheme}, we used a public tool called Morfessor \cite{morfessor}, which can split a word into morphological segments with prefix, stem, and suffix tags. For \emph{Syllable}, we implemented the hyphenation tool proposed by Liang \cite{Liang83wordhy-phen-a-tion}, which has been used in many editing softwares like \LaTeX{} to break words by syllables.

For each of them, given a word $w$, we calculated its similarities to all the other words and selected the top 5 words with highest similarity to build the relation edges in the weight matrix $R$.\footnote{In our experiments, the performance varies little when the number of similar words varies from 3 to 50.} We tested the $R$ matrix built based on each single type of knowledge, and we also tested the $R$ matrix built based on several types of knowledge through combination. Specifically, given the four ranked lists of words from the morphological knowledge, we combined them into a union set, and selected the top 5 words that got more votes by the four knowledge types.

\subsubsection{The Balancing Parameters}
As discussed in Section \ref{sec-model}, the balancing parameters in KNET might be related to word frequency. For simplicity, we used a common algorithm to divide the words into a certain number of buckets. Specifically, suppose that we want to have $b$ buckets, then we rank the words in the vocabulary by their frequencies in the descending order, and put the words into the first bucket one by one until the summed frequency of the first bucket reaches the $1/b$ of the total word frequency. After that, we feed the rest buckets in the similar way, and eventually the summed frequency of each of the $b$ buckets is approximately equal to $1/b$ of the total frequency. We let all words in one bucket share the same balance coefficients. In our experiments, we initialize both $c_1$ and $c_2$ with 0.5 in each bucket. We set the number of buckets to 1000 on the analogical reasoning task and WordSim-353 word similarity task since they are mainly composed of frequent words, while we set the number of buckets to 100 thousand on RareWord word similarity task since it contains many rare and unknown words. More discussion about the balancing between contextual information and morphological knowledge as well as the relationship with the number of buckets can be found in Section \ref{sec-analysis}.

\subsection{Comparison with Baselines Related to Skip-gram}\label{sec-exp_skip-gram}

\subsubsection{Datasets}

The training set used in this part of experiments is the \emph{enwik9} data\footnote{http://mattmahoney.net/dc/enwik9.zip}, which is built from the first billion characters from Wikipedia. This corpus contains totally 123.4 million words. We used Matt Mahoney's text pre-processing script\footnote{http://mattmahoney.net/dc/textdata.html} to process the corpus. After pre-processing, all digits were replaced with English words (e.g., \emph{3} was replaced with \emph{three}), and the metadata and hyperlinks were removed. Furthermore, all words that occurred less than 5 times in the training data were discarded from the vocabulary, resulting in a vocabulary of 220 thousand words. The out-of-vocabulary words were ignored in training.

\subsubsection{Compared Methods and Experimental Settings}\label{sec-exp_setting}

In our experiments, we compare the following methods:

\textbf{Skip-gram}: this is a popularly used baseline model introduced by \cite{Mikolov2013nips}.

\textbf{Skip-gram + Edit/LCS/Morpheme/Syllable/Combination Input Feature}: this is another group of baselines using the morphological features as additional inputs during training of the Skip-gram model. Specifically, the input is no longer a $1$-of-$V$ representation. Instead, we will append the morphological feature, which is the corresponding row of the relation matrix $R$, to the $1$-of-$V$ vector. Thus the input is a vector of length $2V$ and the projection matrix has the size of $2V\times D$ where $D$ is the dimension of word embeddings. We denote this group of baselines as Skip-gram + Input Feature.

\textbf{Skip-gram + Fixed Edit/LCS/Morpheme/Syllable/Combination Relation Matrix}: this is the same with our proposed model except that we do not update the relation matrix while learning the word embedding. We design this baseline to verify that blindly sticking to the morphological knowledge may even hurt in some cases which is coherent with the human cognitive psychology. We denote this group of baselines as Skip-gram + Fixed Relation Matrix.

\textbf{Skip-gram + Edit/LCS/Morpheme/Syllable/Combination Relation Matrix}: this is the proposed KNET model, in which we employed the same types of morphological knowledge and update all the parameters in the training process. Note that, our model can be degraded to the second baseline by fixing $c_1$, $c_2$, $R$ and not sharing $M$. If we only fix $R$ in KNET, we get the third baseline.

In all the above methods, we set the dimension of word embeddings to 100 and the context windows size to 5. We employed the negative sampling technique to train these models and the number of negative samples was set to 3.

With the above settings, the training time of the proposed model was only about 1.5 times of the original Skip-gram model, showing that the KNET framework is very efficient. Actually its training can finish in about 15 minutes on a single machine with four cores.

\subsubsection{Results}

Table \ref{tab:baseline} shows the performance of the methods on the two tasks, respectively. RareWord (known words) shows the results obtained by representing all unknown words as a default vector. RareWord (all words) shows the results obtained by predicting the embedding of unknown words with the relation matrix and the embedding of known words using the method described in Section \ref{sec:wordSimTask}. In the vertical direction, we can find that the performance of model groups follows the order of \textbf{Skip + Input Feature $\prec$ Skip-gram $\prec$ Skip-gram + Fixed Relation Matrix $\prec$ Skip-gram + Relation Matrix (KNET)}, where $\prec$ means worse than.
\smallskip

\begin{table*}
\centering\vspace{0pt}
\tbl{Comparison between KNET and baselines related to Skip-gram on analogical reasoning task and word similarity task. We report the semantic/syntactic/total accuracy in analogical reasoning task and Spearman's $\rho$ correlation in word similarity task. The word embeddings are trained on \emph{enwiki9} data with dimension 100.\label{tab:baseline}}{
\begin{tabular}{|l|c|c|c|c|c|c|}
\hline
 & \multicolumn{3}{|c|}{\textbf{Analogical Reasoning Task}}  & \multicolumn{3}{|c|}{\textbf{Word Similarity Task}} \\
\hline
 Model &   Semantic &  Syntactic &      Total & WordSim-353 &   RareWord & RareWord\\
 & Accuracy & Accuracy  & Accuracy & & (known words) & (all words) \\
\hline

Skip-gram &    21.85\% &    34.64\% &    28.84\% &     0.6283 &     0.1685 &     -       \\
\hline
+ Edit Input Feature &    13.67\% &    27.85\% &    21.41\% &     0.5788 &     0.1625 &     0.3087       \\

+ LCS Input Feature &    13.65\% &    28.30\% &    21.65\% &     0.6055 &     0.1679 &      0.3180      \\

+ Morpheme Input Feature &    13.55\% &    23.66\% &    19.07\% &     0.5954 &     0.1595 &      0.3068      \\

+ Syllable Input Feature &    11.94\% &    25.30\% &    19.24\% &     0.5657 &     0.1554 &       0.2944     \\

+ Combination Input Feature &    13.67\% &    28.52\% &    21.78\% &     0.5759 &     0.1659 &       0.3228     \\
\hline
+ Fixed Edit Relation Matrix &    21.42\% &    40.62\% &    31.91\% &      0.6384      &     0.1962       &      0.3595      \\

+ Fixed LCS Relation Matrix &    23.48\% &    41.24\% &    33.18\% &     0.6452       &      0.1982      &    0.3609        \\

+ Fixed Morpheme Relation Matrix &    23.94\% &    41.04\% &    33.28\% &     0.6451    &   0.1800    &   0.3235       \\

+ Fixed Syllable Relation Matrix &    22.48\% &    40.61\% &    32.38\% &    0.6482    &   0.1814   &    0.3301        \\

+ Fixed Combination Relation Matrix &    21.17\% &    43.60\% &    33.42\% &   0.6423     &   0.2085    &    0.3686      \\
\hline

+ Edit Relation Matrix &    23.59\% &    43.49\% &    34.46\% &     0.6532 &     0.2103 &   0.3797 \\

+ LCS Relation Matrix &    23.70\% &    44.50\% &    35.06\% &     0.6545 &     0.2043 &   0.3700 \\

+ Morpheme Relation Matrix &    \textbf{24.86\%} &    43.68\% &    35.14\% &     \textbf{0.6612} &     0.1909 &   0.3347 \\

+ Syllable Relation Matrix &    24.53\% &    41.88\% &    34.01\% &     0.6607 &     0.1916 &   0.3371 \\

+ Combination Relation Matrix &    23.58\% &    \textbf{46.90\%} &   \textbf{ 36.32\%} &     0.6495 &    \textbf{0.2191} &   \textbf{0.3932} \\
\hline
\end{tabular}
}
\end{table*}

By \textbf{Skip-gram $\prec$ Skip-gram + Fixed Relation Matrix} and \textbf{Skip-gram $\prec$ Skip-gram + Relation Matrix (KNET)}, we can observe that
\begin{enumerate}
	\item Adding morphological knowledge, either single type or combined knowledge, to the Skip-gram model can consistently increase all types of accuracies in the analogical reasoning task and word similarity task. This shows that morphological knowledge can effectively improve the quality of the learned word embeddings.
	\item Looking into the inside of Skip-gram + Relation Matrix (KNET) group, we can find that \emph{Morpheme} performs the best among the five types of knowledge in terms of semantic accuracy in analogical reasoning task and WordSim-353 in word similarity task. Since these two tasks focus on semantic relationship, we hypothesize the reason is that morphemes (like roots and affixes) are basic units in word composition, and it implies accurate semantic correlation if two words share the same root. On the other hand, \emph{Combination} performs the best in terms of syntactic accuracy in the analogical reasoning task and RareWord in the word similarity task. Besides, \emph{Edit} and \emph{LCS} are always better than \emph{Morpheme} and \emph{Syllable} in these two tasks. The possible reason is that in these tasks the recall of the truly similar words is more critical than the precision. \emph{Edit} and \emph{LCS} naturally have high recall of the truly similar words because they directly calculate the similarity score in the letter level. Even though \emph{Edit} and \emph{LCS} have high recall, every single type of morphological knowledge has its own limitations, and thus combining them together will further increase the recall of truly similar words which leads to better performance on these two tasks.
	\item Focus on the performance of Skip-gram + Relation Matrix (KNET) on word similarity task, we can have the following observations. By using the embeddings of known words and relation matrix to predict those of the unknown words, we can achieve significant improvement on the RareWord set, with almost 100\% increment compared with the baseline methods. This indicates that our proposed KNET framework can effectively deal with new emerging words, which yields potential impact for natural language processing applications in real world. The average gain on RareWord (10.08\%) is much higher than that on WordSim-353 (4.38\%), which illustrates that leveraging morphological knowledge will especially benefit rare words. Since there is no sufficient context information for the rare words in the training data, building connections between words using the morphological knowledge will provide additional evidence for us to generate effective embeddings for these rare words. While the rare words can benefit from the morphological knowledge, we can keep the noise brought by it away from the frequent words. The secret is that in our KNET framework, frequent words rely more on the contextual information while rare words rely more on the morphological knowledge by balancing between these two branches. More discussion can be seen in Section \ref{sec-analysis}.
\end{enumerate}

By \textbf{Skip-gram + Input Features $\prec$ Skip-gram $\prec$ Skip-gram + Fixed Relation Matrix}, we can observe that
simply adding morphological knowledge as additional input features does not work as expected and conversely hurt the language model. Recall that our KNET framework can be degraded to Skip-gram + Input Features by fixing $c_1$, $c_2$, $R$ and not sharing $M$ in the training process. We can also obtain Skip-gram + Fixed Relation Matrix by fixing $R$ from KNET. Thus the difference between Skip-gram + Input Features and Skip-gram + Fixed Relation Matrix is fixing $c_1$, $c_2$ and not sharing $M$, which leads to the great gap of performance, i.e., one is worse than Skip-gram and the other is better than it. It indicates that $c_1$, $c_2$ and sharing $M$ brings the core effectiveness of our proposed KNET framework. Actually $M$ is the channel that the contextual information branch and the morphological branch used to communicate with each other, while $c_1$ and $c_2$ are the key factors of balancing between these two branches. With both the above two aspects, our KNET framework can effectively leverage the morphological knowledge while keeping consistent with the context. In this perspective, we can easily understand that the language model suffers from the artificially appended identification when we simply add the noisy morphological knowledge as additional input features.

By \textbf{Skip-gram + Fixed Relation Matrix $\prec$ Skip-gram + Relation Matrix (KNET)}, we verified the hypothesis that
blindly sticking to the morphological knowledge may even hurt in some cases and we can leverage the context to avoid the misrecognition brought by the morphological knowledge which is coherent with the human cognitive psychology introduced in Section \ref{sec-human}. In this manner, the contextual information branch and the morphological knowledge branch can   reinforce each other. The morphological knowledge helps when the context is insufficient while the context can correct and refine the noisy morphological knowledge.

To sum up, through the comparison of experiment results among these models, we can claim that our KNET framework is a general, effective, and robust framework that can leverage both contextual information and morphological knowledge while making them harmonize with each other. Specifically, with sharing $M$ these two branches can communicate, with updating $c_1$, $c_2$ these two branches can balance, and with updating $R$, $M$ these two branches can reinforce each other as a united framework. By analyzing the results of KNET on two word similarity tasks, we find that our framework can learn the effective word embedding especially on rare words, which will be further verified in the next subsection.

\subsection{Comparison with the morphoRNN model}\label{sec:morphoRNN}
\subsubsection{Datasets and Experimental Settings}

To make the comparison fair, we used the same dataset as \cite{luong2013better}, which is the April 2010 snapshot of the Wikipedia corpus denoted as \emph{wiki2010}. After the pre-processing similar to \emph{enwiki9} and ignoring the words that occurred less than 10 times, there are 487 million tokens with a vocabulary of 466 thousand words. The experimental settings are almost the same with the previous experiments except that we set the dimension of word embeddings to 50 to be consistent with \cite{luong2013better}. Because they didn't publish the codes and only published the trained word embedding on \emph{wiki2010}, considering there are many words in the analogical reasoning task but not in the vocabulary of \emph{wiki2010}, we can not fairly compare morphoRNN with KNET on the analogical reasoning task. Therefore we will focus on the word similarity task which is also the main part in their work, and we will refer the numbers reported in their paper.

\subsubsection{Compared Methods and Results}
The morphoRNN was proposed by Luong \emph{et al} \cite{luong2013better}, which has been introduced in Section \ref{sec-related}.  In their work, they proposed two kinds of morphoRNNs, cimRNN which is context insensitive and  csmRNN which is context sensitive. Since the context sensitive models are consistently better than the context insensitive models as expected, we only compare KNET with their context sensitive models. In their experiments, they make use of two publicly-available embeddings provided by \cite{Collobert2011} and \cite{HuangEtAl2012} to initialize their models. Following their notation, we denote these two morphoRNN models as \textbf{C\&W + csmRNN} and \textbf{HSMN + csmRNN} which are the best models in their work. The results of these two models and our KNET models are shown in Table \ref{tab:morphoRNN}.

\begin{table*}
\centering\vspace{0pt}
\tbl{Comparison between KNET and morphoRNN on word similarity task measured by the Spearman's $\rho$ correlation. The word embeddings are trained on \emph{wiki2010} data with dimension 50. Here we refer the numbers reported in their paper directly.\label{tab:morphoRNN}}{
\begin{tabular}{|l|c|c|}
\hline
  Model         & WordSim-353 &   RareWord \\
\hline

 HSMN + csmRNN &     \textbf{0.6458} &     0.2231 \\
C\&W + csmRNN &     0.5701 &     0.3436 \\
\hline
 Skip-gram &     0.6010 &     0.2855 \\
\hline
 + Edit Relation Matrix &     0.5953 &     0.3714 \\

 + LCS Relation Matrix &     0.6076 &     \textbf{0.3780} \\

 + Morpheme Relation Matrix &     0.5983 &     0.3647 \\

 + Syllable Relation Matrix &     0.6021 &     0.3715 \\

 + Combination Relation Matrix &     0.6094 &     0.3752 \\
\hline
\end{tabular}
}
\vspace{0pt}
\end{table*}

From the results we can observe that the best model on WordSim-353 is HSMN + csmRNN. As explained in  \cite{luong2013better}, the reason is that HSMN performs good on frequent words and HSMN + csmRNN uses the word embedding produced by HSMN in the initialization. However, although HSMN + csmRNN can do a great job on frequent words, its performance on RareWord is bad. Although C\&W + csmRNN performs better than HSMN + csmRNN, they are both beaten by the proposed KNET models powered by different types of morphological knowledge, showing the effectiveness of KNET. Compared with Skip-gram, our proposed model can greatly improve the performance on RareWord while the performance on WordSim-353 is flat. This is reasonable because the morphological knowledge can help improve the quality of word embeddings for rare words while can barely help words with already plenty context information especially in the low dimension embedding space. Note that the performance of the models in Table \ref{tab:morphoRNN} is a little worse than those in Table \ref{tab:baseline} which is because the dimension of word embeddings is 50 on \emph{wiki2010} while it is 100 on \emph{enwiki9}.

Besides the promising performance on rare words, KNET has several other advantages over morphoRNN.
\begin{enumerate}
  \item KNET is much more efficient, since it does not need initialization by other word embeddings. In contrast, C\&W + csmRNN is initialized with the C\&W embeddings which were trained for about 2 months. Furthermore, KNET is much more efficient than morphoRNN models in both the language model and the recursive structure, so that it can be trained in less than 20 minutes on a single machine with four cores.
  \item KNET is more robust, since it can benefit from the noisy knowledge by updating the relation matrix and balancing between contextual information and morphological knowledge. In contrast, morphoRNN models used a hierarchical structure to co-consider the morphological knowledge and the contextual information, and thus the noise accumulated in the morphological layer (the RNN structure) might be propagated to the context layer (the language model).
  \item KNET is more flexible, since it can not only leverage the morpheme knowledge but also other morphological knowledge types such as \emph{Edit} and \emph{LCS}, which is not applicable for morphoRNN. Actually, KNET can leverage any kind of pairwise relationship which can cover most of the relations in knowledge bases such as \emph{WordNet} and \emph{Freebase}. We leave this for future work.
\end{enumerate}

\begin{table*}[h]
  \centering\vspace{0pt}
  \tbl{Top five similar words in the embedding spaces produced by KNET using the combination of morphological knowledge.\label{table:examples}}{
    \begin{tabular}{|c|c|c|c|}
    \hline
    Example Word & Skip-gram & Combined Knowledge & Skip-gram + Combination Relation Matrix \\
    \hline
    \hline
    \multirow{5}[0]{*}{uninformative} & monotherapy & informative & problematic \\
     &  lcg  & inchoative & fallacious \\
     & electrodeposition & inoperative & inaccurate \\
     & astrophotography & interrogative & uninteresting \\
     & ultrafilters & formative & precisely \\
     \hline
    \multirow{5}[0]{*}{stepdaughter} & grandaughter & daughters & grandaughter \\
     & swynford & daughter & daughter \\
     & caesaris & grandaughter & daughters \\
     & theling & steptoe & wife \\
     & stepson & slaughter & stepfather \\
     \hline
    \multirow{5}[0]{*}{uncompetitive} & overvalued & competitively & competitive \\
     & monopsony & competitive & noncompetitive \\
     & skyrocketing & noncompetitive & profitable \\
     & dampened & competitiveness & competetive \\
     & undervalued & competetive & lucrative \\
     \hline
    \multirow{5}[0]{*}{tasteful} & hackneyed & wasteful & tastes \\
     & freshest & distasteful & piquant \\
     & haircuts & tasted & pretentious \\
     & nutritive & distaste & taste \\
     & teapots & tastes & elegance \\
    \hline
    \multirow{5}[0]{*}{weirdest} & swordfight & weird & weird \\
     & merseybeat & weirdos & fun \\
     & sty   & widest & nostalghia \\
     & oversoul & wildest & weirdos \\
     & washroom & nordeste & skinflint \\
    \hline
    \end{tabular}\vspace{0pt}
  }
\end{table*}

\subsection{Case Study}

To further understand that how KNET benefits from the noisy morphological knowledge, we sampled some rare words and compare the closest words to them in different word embedding spaces and morphological knowledge to check the effect of learning process. Specifically, for a given word, we extracted its representation vector in the 100-dimension embedding space which we obtained in Section \ref{sec-exp_skip-gram}, and calculated its cosine similarity with the representation vectors of all the other words. Then we show the five most similar words generated by the methods under investigation in Table \ref{table:examples}. According to Table \ref{tab:baseline}, the combination of four types of knowledge achieved the best performance on most tasks, therefore we only show the results for the baseline method (Skip-gram) and the combination method (Skip-gram + Combination Relation Matrix). Besides, we also show the most similar words directly given by the combination of the four types of knowledge without going through the learning process (denoted as Combined Knowledge), which can give us an overview of how the original morphological knowledge looks like. Note that actually the baseline does a good job on frequent words and the results of our model on those words are similar to the baseline, so we only sampled some rare words to demonstrate the power of the KNET model.
\smallskip

From Table \ref{table:examples}, we have the following observations:
\begin{enumerate}
  \item We can see that the Skip-gram method often fails in finding reasonable semantically or syntactically related words for rare words. For example, \emph{uninformative} only appears 18 times in the training corpus, and thus its nearest neighbors are almost random. According to the morphological knowledge (see the column of Combined Knowledge), this word may have relation with \emph{informative} and \emph{formative}. By leveraging these relatively frequent words to enhance the embedding for \emph{uninformative}, our model eventually generate very effective embedding for this rare word, and its similar words in the learned embedding space become much more reasonable.
  \item We can also see that the morphological knowledge could be noisy in some cases. For example, it suggests \emph{inchoative} and \emph{interrogative} to \emph{uninformative}, because these words share a substring \emph{ative} with \emph{uninformative}. However, they are neither syntactically similar nor semantically similar. The power of our proposed framework lies in that it can distinguish useful knowledge and noise by seeking help from the contextual information, and refine the tradeoff coefficients and the relationship matrix to ensure the generation of a more reliable embedding. We can see that the most similar words to \emph{uninformative} in the final embedding space, such as \emph{problematic} and \emph{inaccurate} are more semantically correlated to \emph{uninformative} than \emph{inchoative} and \emph{interrogative}.
\end{enumerate}

To sum up, the examples in Table \ref{table:examples} indicate that for rare words, (i) it is unreliable to learn their embeddings only from contexts; (ii) morphological knowledge can do a great favor if we can successfully deal with the noise it brings in; (iii) contextual information can help in distinguishing useful knowledge and noise. In this manner, our proposed KNET framework can achieve the best performance while the contextual information and morphological knowledge harmonize with each other.

\subsection{Analysis of the Balancing Function}\label{sec-analysis}

In this subsection, we give some empirical results on the influence of the balancing function between the contextual information branch and the morphological knowledge branch in the KNET framework. The greater the ratio of the tradeoff coefficients (i.e., $c_1/c_2$) is, the more the model relies on the contextual information branch. By analyzing the variation of this ratio under different settings, we can draw the following two conclusions:
\begin{enumerate}
  \item For a specific model, frequent words rely more on contextual information while rare words rely more on morphological knowledge.
  \item By comparing the overall weighted ratios of different models under different settings, we can observe that: (i) models relying more on contextual information perform better than those relying more on morphological knowledge on tasks mainly composed of frequent words; (ii) models relying more on morphological knowledge perform better than those relying more on contextual information on tasks mainly composed of rare words.
\end{enumerate}

We give more detailed discussions about the two conclusions as below.

\subsubsection{Rare Words Rely More on Morphological Knowledge}

\begin{figure}[ht]
\begin{center}\vspace{0pt}
\centerline{\includegraphics[width=1.2\columnwidth]{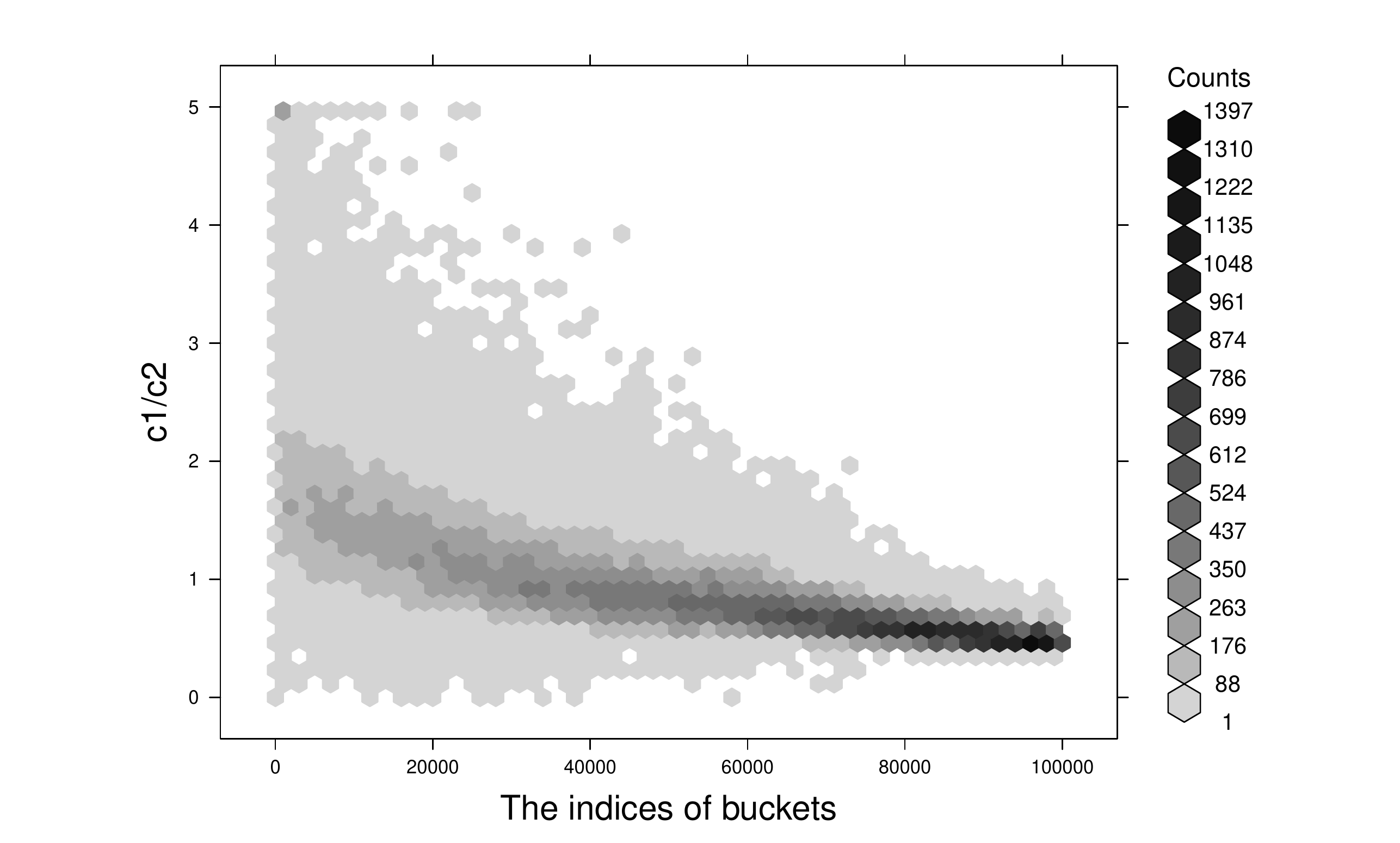}}
\caption{The ratio of $c_1/c_2$ in different buckets}
\label{fig:c1_c2}
\end{center}
\vskip 0in
\end{figure}

We use the results of Skip-gram + Combination Relation Matrix to illustrate the first conclusion. Note that we observed the similar phenomenon for other models. As we want to carefully analyze the behaviour of words with different frequencies, we set the number of buckets $b$ to 100 thousand so that each bucket may only averagely contain two words. The hexagon binning plot of ratios $c_1/c_2$ in different buckets is shown in Figure \ref{fig:c1_c2}, in which the indices of the buckets are in the descending order according to the frequency, i.e., the first bucket contains the most frequent words and the last bucket contains the rarest words. We took the absolute value for each ratio and fix the ratios greater than 5 to be 5 so as to make the figure more readable. The gray scale of the hexagon represents the number of points falling in that
hexagon, i.e., the hexagon is darker when more points fall in it.

We can see in Figure \ref{fig:c1_c2} that the ratio $c_1/c_2$ is approximately decreasing as the indices of the buckets increases, which indicates that frequent words have relatively higher ratios of $c_1/c_2$ than rare words. In other words, frequent words rely more on contextual information and rare words rely more on morphological knowledge. Recall that we initialize both $c_1$ and $c_2$ of each bucket with the same value 0.5, it is surprising that the model learns how to leverage morphological knowledge according to word frequency all by itself.

\subsubsection{Models Relying More on Morphological Knowledge Perform Better on Rare Words}

\begin{figure}[ht]
\begin{center}\vspace{0pt}
\centerline{\includegraphics[width=1.0\columnwidth]{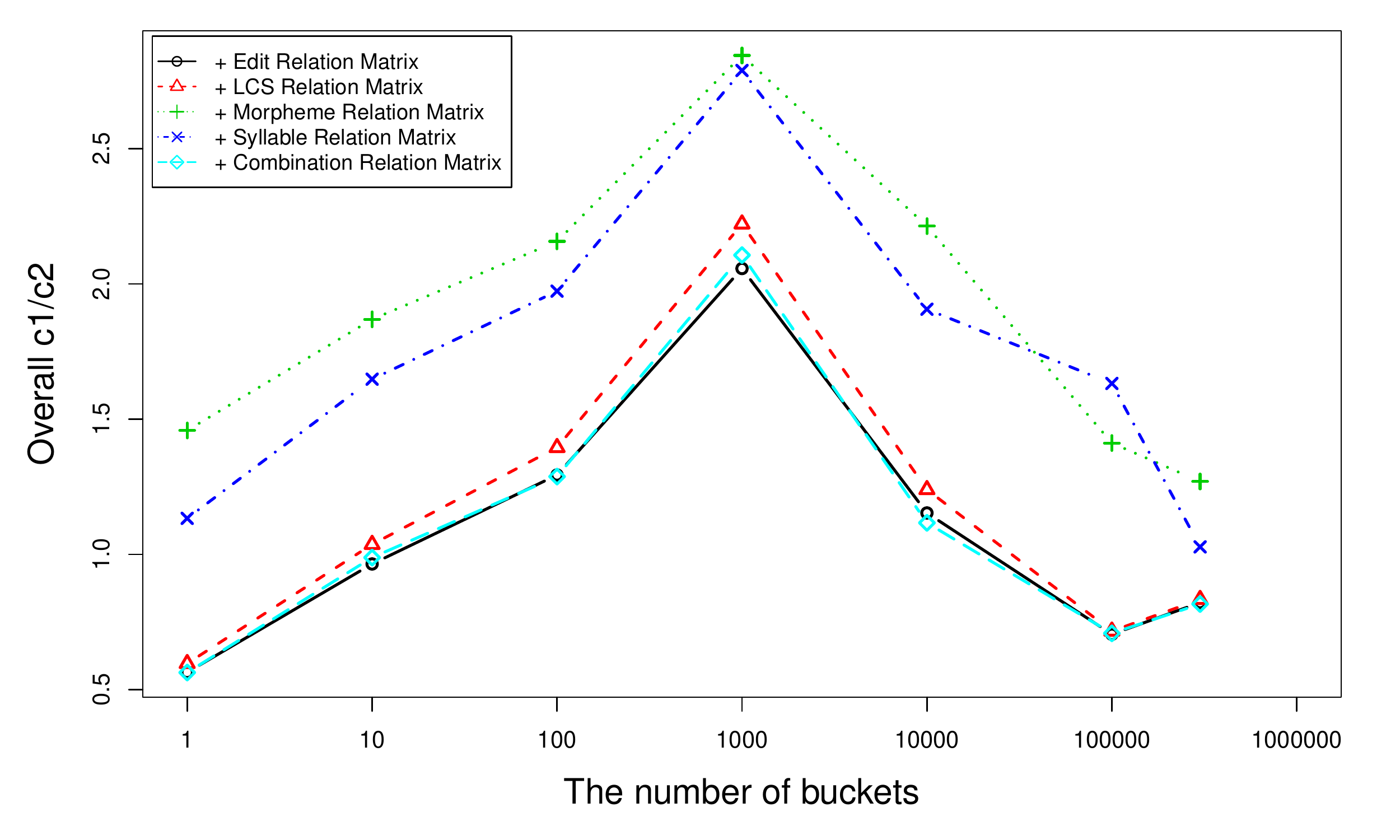}}
\caption{The overall ratios of $c_1/c_2$ in different models while the number of buckets varies.} \vspace{0pt}
\label{fig:overall_c1_c2}
\end{center}
\vskip 0in
\end{figure}

\begin{figure}[ht]
\begin{center}\vspace{0pt}
\centerline{\includegraphics[width=1.0\columnwidth]{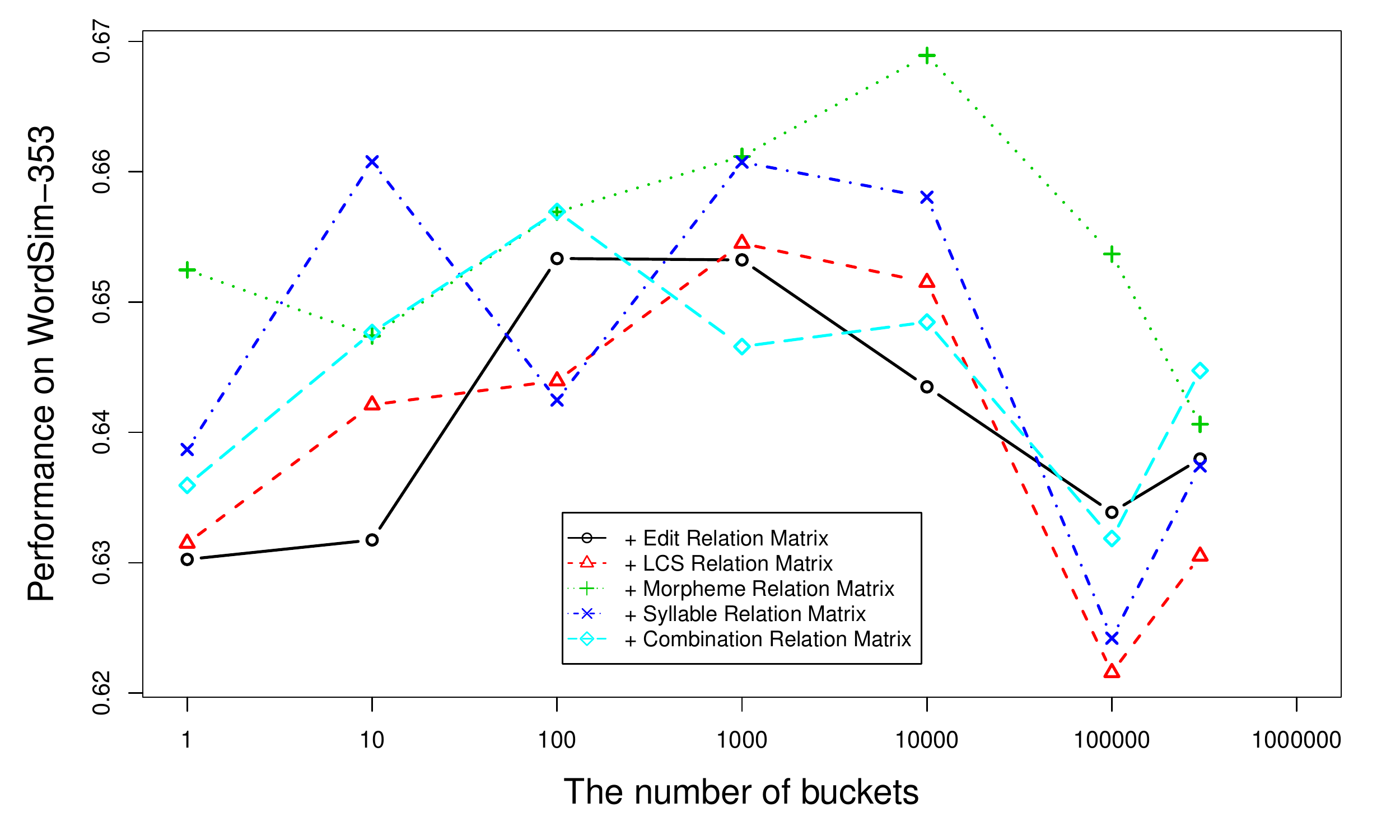}}
\caption{The performance of different models on WordSim-353 while the number of buckets varies.} \vspace{0pt}
\label{fig:wordsim-353}
\end{center}
\vskip 0in
\end{figure}

\begin{figure}[ht]
\begin{center}\vspace{0pt}
\centerline{\includegraphics[width=1.0\columnwidth]{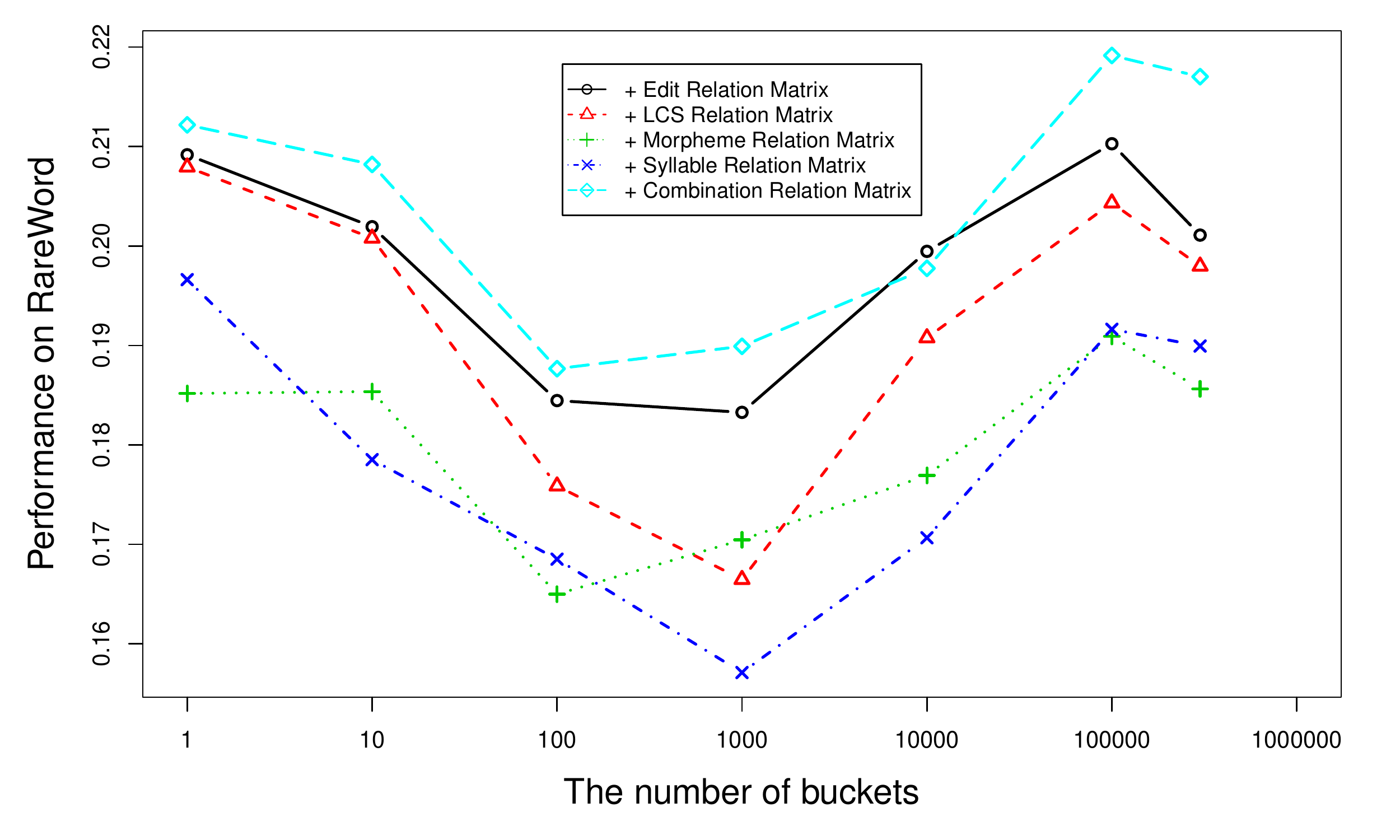}}
\caption{The performance of different models on RareWord while the number of buckets varies.} \vspace{0pt}
\label{fig:rareword}
\end{center}
\vskip 0in
\end{figure}

We compare all the proposed knowledge-powered word embedding models under different settings to illustrate the second conclusion. Specifically, we evaluated Skip-gram + Edit/LCS/Morpheme/Syllable/Combination Relation Matrix that were trained with different numbers of buckets on WordSim-353 and RareWord. As WordSim-353 mainly contains frequent words and RareWord contains many rare words, we can analyze the results on these two datasets to estimate the performance on frequent words and rare words. The overall ratio of each model is computed as the weighted sum of $c_1$ over all buckets divided by the weighted sum of $c_2$ over all buckets like below,
$$c_1/c_2=\frac{\sum_{i=1}^b c_{1i}n_i}{\sum_{i=1}^b c_{2i}n_i},$$
where $n_i$ is the number of words in the $i$-th bucket.

The overall ratio of $c_1/c_2$ of different models while the number of buckets varies is shown in Figure \ref{fig:overall_c1_c2}.\footnote{We only run experiments when the number of buckets is 1, 10, 100, 1000, 10000, 100000, 300000 (the maximum value) and connect the points in the figure.} We can see that the overall ratio is the largest when the number of buckets lies in the middle while it drops down as the number of buckets moves to the two extremes. Besides, we can observe that the ratios of Skip-gram + Morpheme/Syllable Relation Matrix are consistently larger than those of Skip-gram + Edit/LCS/Combination Relation Matrix.

The performance of these models on WordSim-353 is shown in Figure \ref{fig:wordsim-353}. The trend is not stable probably due to the uncertainty brought by the small size of WordSim-353. However, we can roughly draw the conclusion that: (i) most models perform better when the number of buckets lies in the middle than in the extremes; (ii) Skip-gram + Morpheme/Syllable Relation Matrix perform better than other models in most cases. The trend of the model performance on WordSim-353 is consistent with the overall ratio, which implies that models relying more on contextual information perform better on frequent words.

The performance of these models on RareWord is shown in Figure \ref{fig:rareword}. The trend is very stable and we can easily observe that it is strictly opposite to the overall ratio of $c_1/c_2$. The models achieve the best performance when the number of buckets is in the two extremes, while the performance is the worst when the number of buckets is in the middle. Besides, the performance of Skip-gram + Morpheme/Syllable Relation Matrix is always worse than that of Skip-gram + Edit/LCS/Combination Relation Matrix. To sum up, we can conclude that models relying more on morphological knowledge perform better on rare words. Note that the points that the number of buckets is the maximal (which means there is only one word in each bucket) look a little strange. The reason is that the $c_1$ and $c_2$ of rare words have very little opportunity to be updated when the buckets are very sparse. Further considering that the initializations of $c_1$ and $c_2$ are both 0.5 which result in the middle values of the ratio $c_1/c_2$, we can understand why the corresponding model performance tends to be in the middle.

\section{Conclusions and Future Work}\label{sec-conclusion}

We proposed a novel neural network framework called KNET to leverage morphological word similarity to learn high-quality word embeddings. The framework contains a contextual information branch to leverage word co-occurrence information and a morphological knowledge branch to leverage morphological relationship between words. We tested the framework on several tasks and the results show that it can produce enhanced word representations compared with the state-of-the-art models.

The proposed KNET framework is also applicable to others morphologically complex languages such as Finnish or Turkish especially when the amount of text data is limited and the vocabularies are huge. For the future work, we plan to leverage other types of relationships (e.g., the relationships in the knowledge bases like \emph{WordNet} and \emph{Freebase}) in the KNET framework to check whether we can obtain even better word representations.

\bibliographystyle{ACM-Reference-Format-Journals}
\bibliography{morphological}


\begin{thebibliography}{00}


\ifx \showCODEN    \undefined \def \showCODEN     #1{\unskip}     \fi
\ifx \showDOI      \undefined \def \showDOI       #1{{\tt DOI:}\penalty0{#1}\ }
  \fi
\ifx \showISBNx    \undefined \def \showISBNx     #1{\unskip}     \fi
\ifx \showISBNxiii \undefined \def \showISBNxiii  #1{\unskip}     \fi
\ifx \showISSN     \undefined \def \showISSN      #1{\unskip}     \fi
\ifx \showLCCN     \undefined \def \showLCCN      #1{\unskip}     \fi
\ifx \shownote     \undefined \def \shownote      #1{#1}          \fi
\ifx \showarticletitle \undefined \def \showarticletitle #1{#1}   \fi
\ifx \showURL      \undefined \def \showURL       #1{#1}          \fi

\bibitem[\protect\citeauthoryear{Bengio and Senecal}{Bengio and
  Senecal}{2003}]%
        {Bengio03ImportanceSampling}
{Y. Bengio} {and} {J.-S. Senecal}. 2003.
\newblock Quick Training of Probabilistic Neural Nets by Importance Sampling.
\newblock   (2003).
\newblock


\bibitem[\protect\citeauthoryear{Bengio and Senecal}{Bengio and
  Senecal}{2008}]%
        {Bengio08ImportanceSampling}
{Y. Bengio} {and} {J.-S. Senecal}. 2008.
\newblock \showarticletitle{Adaptive Importance Sampling to Accelerate Training
  of a Neural Probabilistic Language Model}.
\newblock {\em Trans. Neur. Netw.\/} {19}, 4 (2008), 713--722.
\newblock


\bibitem[\protect\citeauthoryear{Bian, Gao, and Liu}{Bian
  et~al\mbox{.}}{2014}]%
        {bian2014knn}
{Jiang Bian}, {Bin Gao}, {and} {Tie-Yan Liu}. 2014.
\newblock \showarticletitle{Knowledge-Powered Deep Learning for Word
  Embedding}. In {\em Proc. of {ECML/PKDD}}.
\newblock


\bibitem[\protect\citeauthoryear{Blei, Ng, and Jordan}{Blei
  et~al\mbox{.}}{2003}]%
        {blei:lda}
{D.M. Blei}, {A.Y. Ng}, {and} {M. Jordan}. 2003.
\newblock \showarticletitle{Latent dirichlet allocation}. In {\em Journal of
  machine learning}.
\newblock


\bibitem[\protect\citeauthoryear{Bordes, Weston, Collobert, Bengio,
  et~al\mbox{.}}{Bordes et~al\mbox{.}}{2011}]%
        {bordes2011learning}
{A. Bordes}, {J. Weston}, {R. Collobert}, {Y. Bengio}, {and} {others}. 2011.
\newblock \showarticletitle{Learning Structured Embeddings of Knowledge
  Bases.}. In {\em AAAI}.
\newblock


\bibitem[\protect\citeauthoryear{Chapman}{Chapman}{1998}]%
        {chapman1998language}
{J.~W. Chapman}. 1998.
\newblock \showarticletitle{Language prediction skill, phonological recoding
  ability, and beginning reading}.
\newblock {\em Reading and spelling: Development and disorders\/} (1998), 33.
\newblock


\bibitem[\protect\citeauthoryear{Collobert and Weston}{Collobert and
  Weston}{2008}]%
        {Collobert2008}
{R. Collobert} {and} {J. Weston}. 2008.
\newblock \showarticletitle{A Unified Architecture for Natural Language
  Processing: Deep Neural Networks with Multitask Learning}. In {\em ICML}.
  ACM, New York, NY, USA, 160--167.
\newblock


\bibitem[\protect\citeauthoryear{Collobert, Weston, Bottou, Karlen,
  Kavukcuoglu, and Kuksa}{Collobert et~al\mbox{.}}{2011}]%
        {Collobert2011}
{R. Collobert}, {J. Weston}, {L. Bottou}, {M. Karlen}, {K. Kavukcuoglu}, {and}
  {P. Kuksa}. 2011.
\newblock \showarticletitle{Natural Language Processing (Almost) from Scratch}.
\newblock {\em JMLR\/}  {12} (2011), 2493--2537.
\newblock


\bibitem[\protect\citeauthoryear{Creutz and Lagus}{Creutz and Lagus}{2007}]%
        {morfessor}
{M. Creutz} {and} {K. Lagus}. 2007.
\newblock \showarticletitle{Unsupervised models for Morpheme segmentation and
  morphology learning}.
\newblock {\em TSLP\/} {4}, 1 (January 2007).
\newblock


\bibitem[\protect\citeauthoryear{Deng, He, and Gao}{Deng et~al\mbox{.}}{2013}]%
        {DengHG13}
{L. Deng}, {X. He}, {and} {J. Gao}. 2013.
\newblock \showarticletitle{Deep stacking networks for information retrieval}.
  In {\em ICASSP}. 3153--3157.
\newblock


\bibitem[\protect\citeauthoryear{Ehri}{Ehri}{2005}]%
        {ehri2005learning}
{L.~C. Ehri}. 2005.
\newblock \showarticletitle{Learning to read words: Theory, findings, and
  issues}.
\newblock {\em Scientific Studies of reading\/} {9}, 2 (2005), 167--188.
\newblock


\bibitem[\protect\citeauthoryear{Ehri, Barr, Kamil, Mosenthal, and
  Pearson}{Ehri et~al\mbox{.}}{1991}]%
        {ehri1991development}
{L.~C. Ehri}, {R. Barr}, {M.L. Kamil}, {P. Mosenthal}, {and} {P.D. Pearson}.
  1991.
\newblock \showarticletitle{Development of the ability to read words}.
\newblock {\em Handbook of reading research\/}  {2} (1991), 383--417.
\newblock


\bibitem[\protect\citeauthoryear{Finkelstein, Gabrilovich, Matias, Rivlin,
  Solan, Wolfman, and Ruppin}{Finkelstein et~al\mbox{.}}{2002}]%
        {Finkelstein2001wordsim353}
{L. Finkelstein}, {E. Gabrilovich}, {Y. Matias}, {E. Rivlin}, {Z. Solan}, {G.
  Wolfman}, {and} {E. Ruppin}. 2002.
\newblock \showarticletitle{Placing Search in Context: The Concept Revisited}.
  In {\em ACM Transactions on Information Systems}.
\newblock


\bibitem[\protect\citeauthoryear{Glorot, Bordes, and Bengio}{Glorot
  et~al\mbox{.}}{2011}]%
        {Glorot2011}
{X. Glorot}, {A. Bordes}, {and} {Y. Bengio}. 2011.
\newblock \showarticletitle{Domain adaptation for large-scale sentiment
  classification: A deep learning approach}. In {\em ICML}.
\newblock


\bibitem[\protect\citeauthoryear{Goswami}{Goswami}{1986}]%
        {goswami1986children}
{U. Goswami}. 1986.
\newblock \showarticletitle{Children's use of analogy in learning to read: A
  developmental study}.
\newblock {\em Journal of Experimental Child Psychology\/} {42}, 1 (1986),
  73--83.
\newblock


\bibitem[\protect\citeauthoryear{Gutmann and Hyv\"{a}rinen}{Gutmann and
  Hyv\"{a}rinen}{2012}]%
        {Gutmann2012NCE}
{Michael~U. Gutmann} {and} {Aapo Hyv\"{a}rinen}. 2012.
\newblock \showarticletitle{Noise-contrastive Estimation of Unnormalized
  Statistical Models, with Applications to Natural Image Statistics}.
\newblock {\em J. Mach. Learn. Res.\/}  {13} (2012), 307--361.
\newblock


\bibitem[\protect\citeauthoryear{Hinton, McClelland, and Rumelhart}{Hinton
  et~al\mbox{.}}{1986}]%
        {Hinton1986dr}
{G.~E. Hinton}, {J.~L. McClelland}, {and} {D.~E. Rumelhart}. 1986.
\newblock \showarticletitle{Distributed representations}. In {\em Parallel
  distributed processing: Explorations in the microstructure of cognition}. MIT
  Press, 3:1137--1155.
\newblock


\bibitem[\protect\citeauthoryear{Hofmann}{Hofmann}{1999}]%
        {hofmann:plsa}
{T. Hofmann}. 1999.
\newblock \showarticletitle{Probabilistic latent semantic analysis}. In {\em
  Proc. of {UAI}}.
\newblock


\bibitem[\protect\citeauthoryear{Huang, Socher, Manning, and Ng}{Huang
  et~al\mbox{.}}{2012}]%
        {HuangEtAl2012}
{Eric~H. Huang}, {Richard Socher}, {Christopher~D. Manning}, {and} {Andrew~Y.
  Ng}. 2012.
\newblock \showarticletitle{{Improving Word Representations via Global Context
  and Multiple Word Prototypes}}. In {\em Annual Meeting of the Association for
  Computational Linguistics (ACL)}.
\newblock


\bibitem[\protect\citeauthoryear{Liang}{Liang}{1983}]%
        {Liang83wordhy-phen-a-tion}
{F.~M. Liang}. 1983.
\newblock {\em Word Hy-phen-a-tion by Com-put-er}.
\newblock {T}echnical {R}eport.
\newblock


\bibitem[\protect\citeauthoryear{Luong, Socher, and Manning}{Luong
  et~al\mbox{.}}{2013}]%
        {luong2013better}
{M.-T. Luong}, {R. Socher}, {and} {C.~D. Manning}. 2013.
\newblock \showarticletitle{Better word representations with recursive neural
  networks for morphology}.
\newblock {\em CoNLL-2013\/}  {104} (2013).
\newblock


\bibitem[\protect\citeauthoryear{Mikolov}{Mikolov}{2012}]%
        {Mikolov2012phd}
{T. Mikolov}. 2012.
\newblock {\em Statistical Language Models Based on Neural Networks}.
\newblock Ph.D. Dissertation. Brno University of Technology.
\newblock


\bibitem[\protect\citeauthoryear{Mikolov, Chen, Corrado, and Dean}{Mikolov
  et~al\mbox{.}}{2013a}]%
        {Mikolov2013w2v}
{T. Mikolov}, {K. Chen}, {G. Corrado}, {and} {J. Dean}. 2013a.
\newblock \showarticletitle{Efficient Estimation of Word Representations in
  Vector Space} {\em (ICLR '13)}.
\newblock


\bibitem[\protect\citeauthoryear{Mikolov, Sutskever, Chen, Corrado, and
  Dean}{Mikolov et~al\mbox{.}}{2013b}]%
        {Mikolov2013nips}
{T. Mikolov}, {I. Sutskever}, {K. Chen}, {G.~S. Corrado}, {and} {J. Dean}.
  2013b.
\newblock \showarticletitle{Distributed Representations of Words and Phrases
  and their Compositionality.}. In {\em NIPS}. 3111--3119.
\newblock


\bibitem[\protect\citeauthoryear{Mnih and Hinton}{Mnih and Hinton}{2008}]%
        {MnihH08}
{A. Mnih} {and} {G.~E. Hinton}. 2008.
\newblock \showarticletitle{A Scalable Hierarchical Distributed Language
  Model}. In {\em NIPS}. 1081--1088.
\newblock


\bibitem[\protect\citeauthoryear{Mnih and Kavukcuoglu}{Mnih and
  Kavukcuoglu}{2013}]%
        {MnihNIPS2013}
{A. Mnih} {and} {K. Kavukcuoglu}. 2013.
\newblock \showarticletitle{Learning word embeddings efficiently with
  noise-contrastive estimation}.
\newblock In {\em NIPS}. 2265--2273.
\newblock


\bibitem[\protect\citeauthoryear{Mnih and Teh}{Mnih and Teh}{2012}]%
        {MnihICML2012NCE}
{A Mnih} {and} {Y.~W. Teh}. 2012.
\newblock \showarticletitle{A fast and simple algorithm for training neural
  probabilistic language models}. In {\em ICML}. Omnipress, New York, NY, USA,
  1751--1758.
\newblock


\bibitem[\protect\citeauthoryear{Morin and Bengio}{Morin and Bengio}{2005}]%
        {HierarchicalSoftmax}
{F. Morin} {and} {Y. Bengio}. 2005.
\newblock \showarticletitle{Hierarchical probabilistic neural network language
  model}. In {\em AISTATS}. 246--252.
\newblock


\bibitem[\protect\citeauthoryear{Mousa, Kuo, Mangu, and Soltau}{Mousa
  et~al\mbox{.}}{2013}]%
        {mousa2013morph}
{Amr El-Desoky Mousa}, {Hong-Kwang~Jeff Kuo}, {Lidia Mangu}, {and} {Hagen
  Soltau}. 2013.
\newblock \showarticletitle{Morpheme-based feature-rich language models using
  deep neural networks for lvcsr of egyptian arabic}. In {\em Proc. of
  {ICASSP}}.
\newblock


\bibitem[\protect\citeauthoryear{Qiu, Cui, Bian, Gao, and Liu}{Qiu
  et~al\mbox{.}}{2014}]%
        {qiu2014coling}
{Siyu Qiu}, {Qing Cui}, {Jiang Bian}, {Bin Gao}, {and} {Tie-Yan Liu}. 2014.
\newblock \showarticletitle{Co-learning of Word Representations and Morpheme
  Representations}. In {\em Proc. of {COLING}}.
\newblock


\bibitem[\protect\citeauthoryear{Socher, Chen, Manning, and Ng}{Socher
  et~al\mbox{.}}{2013}]%
        {socher2013reasoning}
{R. Socher}, {D. Chen}, {C.~D. Manning}, {and} {A. Ng}. 2013.
\newblock \showarticletitle{Reasoning With Neural Tensor Networks for Knowledge
  Base Completion}. In {\em NIPS}. 926--934.
\newblock


\bibitem[\protect\citeauthoryear{Socher, Lin, Ng, and Manning}{Socher
  et~al\mbox{.}}{2011}]%
        {Socher2011RNN}
{R. Socher}, {C.~C. Lin}, {A.~Y. Ng}, {and} {C.~D. Manning}. 2011.
\newblock \showarticletitle{Parsing Natural Scenes and Natural Language with
  Recursive Neural Networks}. In {\em ICML}.
\newblock


\bibitem[\protect\citeauthoryear{Sperr, Niehues, and Waibel}{Sperr
  et~al\mbox{.}}{2013}]%
        {sperr2013morph}
{Henning Sperr}, {Jan Niehues}, {and} {Alex Waibel}. 2013.
\newblock \showarticletitle{Letter n-gram-based input encoding for continuous
  space language models}. In {\em Proc. of the Workshop on Continuous Vector
  Space Models and their Compositionality}.
\newblock


\bibitem[\protect\citeauthoryear{Turian, Ratinov, and Bengio}{Turian
  et~al\mbox{.}}{2010}]%
        {TurianRB10}
{J.~P. Turian}, {L.-A. Ratinov}, {and} {Y. Bengio}. 2010.
\newblock \showarticletitle{Word Representations: A Simple and General Method
  for Semi-Supervised Learning}. In {\em ACL}. 384--394.
\newblock


\bibitem[\protect\citeauthoryear{Turney}{Turney}{2013}]%
        {Turney:arXiv1310.5042}
{P.~D. Turney}. 2013.
\newblock \showarticletitle{Distributional semantics beyond words: Supervised
  learning of analogy and paraphrase}.
\newblock {\em TACL\/} (2013), 353--366.
\newblock


\bibitem[\protect\citeauthoryear{Turney and Pantel}{Turney and Pantel}{2010}]%
        {Turney2010}
{P.~D. Turney} {and} {P. Pantel}. 2010.
\newblock \showarticletitle{From Frequency to Meaning: Vector Space Models of
  Semantics}.
\newblock {\em Journal of Artificial Intelligence Research\/}  {37} (2010),
  141--188.
\newblock


\bibitem[\protect\citeauthoryear{Weston, Bordes, Yakhnenko, and Usunier}{Weston
  et~al\mbox{.}}{2013}]%
        {weston2013connecting}
{J. Weston}, {A. Bordes}, {O. Yakhnenko}, {and} {N. Usunier}. 2013.
\newblock \showarticletitle{Connecting language and knowledge bases with
  embedding models for relation extraction}.
\newblock {\em arXiv preprint arXiv:1307.7973\/} (2013).
\newblock


\bibitem[\protect\citeauthoryear{Yu and Dredze}{Yu and Dredze}{2014}]%
        {Yu:2014}
{Mo Yu} {and} {Mark Dredze}. 2014.
\newblock \showarticletitle{Improving Lexical Embeddings with Semantic
  Knowledge}. In {\em Association for Computational Linguistics (ACL)}.
\newblock


\end{thebibliography}

\end{document}